# High Performance Human Face Recognition using Independent High Intensity Gabor Wavelet Responses: A Statistical Approach

Arindam Kar[1], Debotosh Bhattacharjee[2], Dipak Kumar Basu[2*], Mita Nasipuri[2], Mahantapas Kundu[2]

[1] Indian Statistical Institute Kolkata-700108, India,
[2] Department of Computer Science and Engineering, Jadavpur University, Kolkata- 700032, India,
* AICTE Emeritus Fellow.
Email: {kgparindamkar@gmail.com, debotosh@indiatimes.com, dipakkbasu@gmail.com, mita_nasipuri@gmail.com, mkundu@cse.jdvu.ac.in}

**Abstract**: In this paper, we present a technique by which high-intensity feature vectors extracted from the Gabor wavelet transformation of frontal face images, is combined together with Independent Component Analysis (ICA) for enhanced face recognition. Firstly, the high-intensity feature vectors are automatically extracted using the local characteristics of each individual face from the Gabor transformed images. Then ICA is applied on these locally extracted high-intensity feature vectors of the facial images to obtain the independent high intensity feature (IHIF) vectors. These IHIF forms the basis of the work. Finally, the image classification is done using these IHIF vectors, which are considered as representatives of the images. The importance behind implementing ICA along with the high-intensity features of Gabor wavelet transformation is twofold. On the one hand, selecting peaks of the Gabor transformed face images exhibit strong characteristics of spatial locality, scale, and orientation selectivity. Thus these images produce salient local features that are most suitable for face recognition. On the other hand, as the ICA employs locally salient features from the high informative facial parts, it reduces redundancy and represents independent features explicitly. These independent features are most useful for subsequent facial discrimination and associative recall. The efficiency of IHIF method is demonstrated by the experiment on frontal facial images dataset, selected from the FERET, FRAV2D, and the ORL database.

## 1. Introduction

Face authentication has gained considerable attention recently, through the increasing need for access verification systems. Such systems are used for the verification of a user's identity on the Internet, when using a bank automaton, when entering to secured building, etc. Face recognition involves computer recognition of personal identity based on geometric or statistical features derived from face images. Even though humans can identify faces with ease, but building such an automated system that accomplishes such objectives is, very challenging. The challenges are even more intense when there are large variations due to illumination conditions, viewing directions or poses, facial expression, aging, etc. The Face recognition research provides the cutting edge technologies in airports, government, military facilities, countries borders, and so on. Principal component analysis (PCA) is a popular statistical method to find useful image representations. Independent Component Analysis (ICA) has emerged over the years as one powerful solution to the problem of blind source separation [1], [2], [3], [4], [5], [6]. Turk and Pentland [7] developed a well-known Eigenfaces method, which sparked great interest in applying statistical techniques to face recognition. While PCA considers the second-order moments only and it un-correlates the data, but ICA would further reduce statistical dependencies and produce a sparse and independent code useful for subsequent pattern discrimination and associative recall [8]. The metric induced by ICA is superior to PCA in the sense that it may provide a representation more robust to the effect of noise [9]. It is, therefore, possible for ICA to be better than PCA for reconstruction in noisy or limited precision environments. When the sources models are sparse, ICA is closely related to the so called non-orthogonal "rotation" methods in PCA and factor analysis. The goal of these rotation methods is to find directions with high concentrations of data. ICA can be used to find interesting non-orthogonal "rotations" [10, 11, 12]. One of the most successful face recognition methods is based on graph matching of coefficients proposed by Lades et. al. [13], which are obtained from Gabor wavelet (GW) responses. However, such graph matching algorithm methods have some disadvantages due to their matching complexity, manual localization of training graphs overall execution time, and extraction of special characteristics of each individual. Bell and Sejnowski [14] has shown that image bases that produce independent outputs from natural scenes are local oriented spatially filters similar to the response properties of simple cells. Conversely, it has been seen that Gabor filters, closely model the responses of simple cells, separate higher-order dependencies [14, 15, 16]. We have already done a work [17], the algorithm proposed for the extraction of high-energized points from Gabor wavelet transforms, has the time complexity of the order $O(N_1 \cdot N_2 \cdot n)$ where N1, N2 is the width and height of the image and n is the number of GW response. If N1=N2= $n$, then the time complexity is of the order $O(n^3)$. The proposed algorithm shown in section III has total computation time $[(5N_1+1)+(5N_1 \cdot (8N_2+1))+5N_1 \cdot 8N_2]$, so the time complexity is $O(N_1 \cdot N_2)$. Thus if N1=N2=$n$, the time complexity is of the order $O(n^2)$. The method introduced in this paper differs from the one in [18], as the latter method integrated the independent properties of only the high-energized feature vectors, which is a nontrivial challenge for implementing fast and automated face recognition systems. As PCA is only sensitive to the power spectrum of images, so it might not be well suited for representing natural images. In particular, it has been observed that images are better described as linear combinations of sources with long tailed distributions [19]. Recent approaches of image representation emphasize on data-driven learning-based techniques, such as the statistical modeling methods [9, 10, 11] the neural network-based learning methods [12], the statistical learning theory and Support vector machine (SVM) based methods [20, 21]. The motivation of using feature based methods is due to their representation of the face image in a very compact way and hence lowering the memory needs.

In this paper, a robust and reliable automatic IHIF method for face recognition is proposed, which is robust to occlusion and illumination changes, and can overcome those disadvantages. This solution is based on selecting intensity peaks of GW responses for the facial high–intensity feature vector construction, instead of using predefined graph nodes as in elastic graph matching (EGM) [13], which reduces representative capability of Gabor wavelets.

The rest of this paper is organized as follows: Section 2 describes Gabor wavelet convolution outputs of facial images. Section 3 describes extraction of high-intensity feature vector from the convolution outputs of Gabor transformed images. Section 4 the application of ICA on the extracted high-intensity feature vector, Section 5 assesses the performance of the IHIF method on the face recognition task by applying it on FERET[22], FRAV2D [23], and the ORL [24] database and comparing with some popular face recognition schemes, and finally we conclude in Section 6.

## 2. 2D Gabor Wavelet Analysis

Physiological studies found simple cells, in human visual cortex, that are selectively tuned to orientation, as well as to spatial frequency, could be approximated by 2D Gabor filters [25, 26, 27]. It has been shown that using Gabor filters as front-end of an automated face recognition system are highly successful [28, 29]. 2D Gabor functions are similar to enhancing edge contours, as well as valleys and ridge contours of the image. The Gabor wavelets, exhibit strong characteristics of spatial locality and orientation selectivity, and are optimally localized in the spatial and frequency domain [26]. The Gabor filter used here has the following general form:

$$\varphi_{\mu,\nu}(z) = \frac{\|k_{\mu,\nu}\|^2}{\sigma^2} e^{-\|k_{\mu,\nu}\|^2 \|z\|^2 / 2\sigma^2} \left[ e^{ik_{\mu,\nu}z} - e^{-(\sigma^2/2)} \right] \quad (1)$$

The center frequency of $i^{th}$ filter is given by the characteristic wave vector

$$\vec{k}_i = \begin{bmatrix} k_{jx} \\ k_{jy} \end{bmatrix} = \begin{bmatrix} k_\nu \cos\theta_\mu \\ k_\nu \sin\theta_\mu \end{bmatrix},$$

having a scale and orientation given by $(k_\nu, \theta_\mu)$, where $\mu$ and $\nu$ defines the orientation and scale of Gabor kernels, $z = (x, y)$, is the variable in spatial domain, $\|\cdot\|$ denotes the norm, and $k_{\mu,\nu}$ is the frequency vector which determines the scale and orientation of Gabor kernels, where $k_\nu = k_{max}/f^\nu$ and $k_{max} = \pi/2$, $\phi_\mu = \pi\mu/8$, $\mu = 0,...7$, where $f$ is the spacing factor between kernels in the frequency domain. Convolving the image with complex Gabor filters with 5 spatial frequency, and eight orientations, captures the whole frequency spectrum, both amplitude and phase $O_{\mu,\nu}(z)$ is denoted as the magnitude of the convolution outputs. The term $e^{-(\sigma^2/2)}$ is subtracted from equation (1) in order to make the kernel DC-free, thus become insensitive to illumination. The decomposition of an image $I$ into these states is called wavelet transformation of the image:

$$R_i(\bar{x}) = \int I(\bar{x}')\varphi_i(\bar{x} - \bar{x}')d\bar{x}' \quad (2)$$

where $I(\bar{x}')$ is the image intensity value at $\bar{x}$, $i = 1, ..., 40$. Fig. 1(a) shows the real part of Gabor kernels, and Fig. 2(b) shows their magnitude with the following parameters : $\sigma = 2\pi$, $k_{max} = \pi/2$, and $f = \sqrt{2}$, at 5 different scales and 8 orientations respectively.

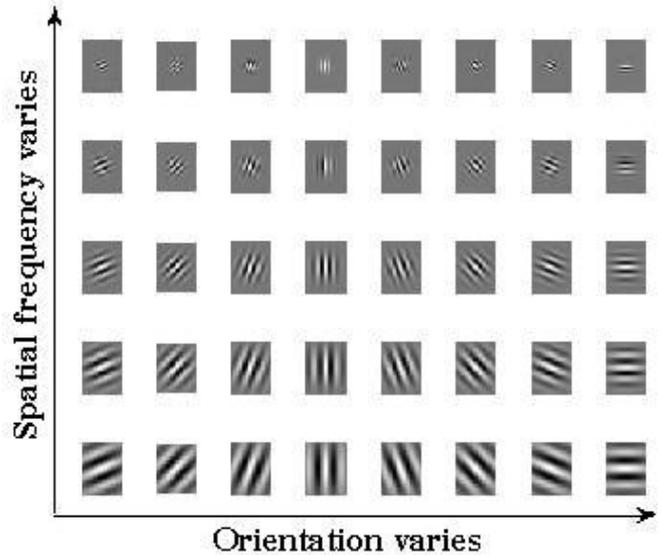

**Fig. 1. (a)** Real part of the Gabor kernels

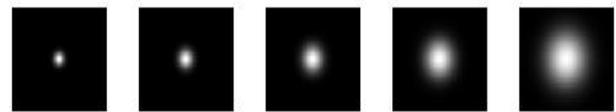

**Fig.1 (b)** Magnitudes of the Gabor kernels at 5 different frequency and 8 orientations

## 3. High-Intensity feature extraction

The Gabor wavelet representation of an image is the convolution of the image with a family of Gabor kernels as defined by (1). Let $I(x,y)$ be the gray level distribution of an image, the convolution output of image I and a Gabor kernel $\varphi_{\mu,\upsilon}$ is defined as follows:

$$O_{\mu,\upsilon}(z) = I(z) * \varphi_{\eta,\upsilon}(z) \quad (3)$$

where $z = (x, y)$, and * denotes the convolution operator. Applying the convolution theorem, we can derive the convolution output from (3) via the fast Fourier transform (FFT)

$$\Im\{O_{\mu,\upsilon}(z)\} = \Im\{I(z)\}\Im\{\varphi_{\mu,\upsilon}(z)\} \quad (4)$$

$$O_{\mu,\upsilon}(z) = \Im^{-1}\{\Im\{I(z)\}\Im\{\varphi_{\mu,\upsilon}(z)\}\} \quad (5)$$

where $\Im$ and $\Im^{-1}$ denote the Fourier and inverse Fourier transform, respectively. The convolution outputs (the magnitude) of a sample image and the Gabor kernels (see fig. 1. (a)) is shown in Fig. 2.

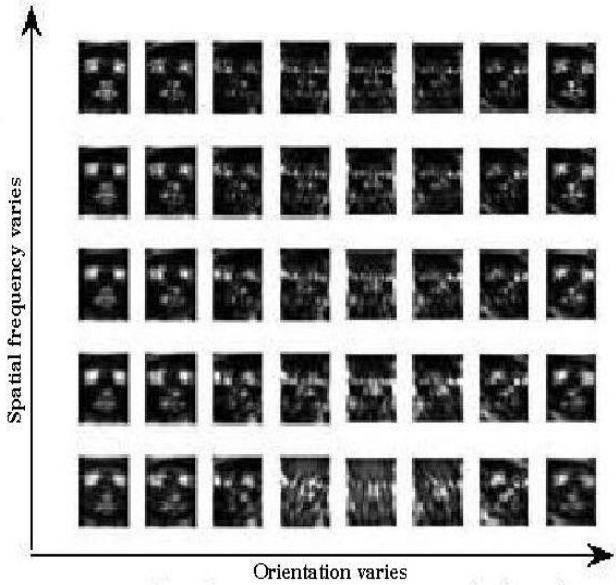

**Fig.2** High intensity Gabor filter responses of a face image

The outputs exhibits strong characteristics of special locality, scale and orientation selectivity, such characteristics produce salient local features such as the eyes, nose, mouth, scars and dimples that are suitable for visual event recognition, and hence making them a suitable choice for feature extraction of images. As the outputs $O_{\mu,\upsilon}(z)$ consist of different local, scale, and orientation features, all these features are concatenated into a single column vector in order to derive a feature vector $V$. The feature extraction of the proposed method has two main steps: (a) High intensity feature point localization (b) Feature vector generation. It is to be noted that we applied the magnitude but did not use the phase, which is considered to be consistent with the application of Gabor representations in [15, 16]. As the outputs $(O_{\mu,\upsilon}(z): \mu \in \{0,...,4\}, \upsilon \in \{0,...7\})$ consists of 40, different local, scale and orientation features, the dimensionality of the Gabor transformed image space is very high. So the following technique is applied for the extraction of low dimensional high intensity vector $X_k$ from the convolution outputs. (a) The algorithm for high intensity feature point localization is as follows:

**Start**

**Step 1**: Find $G_{ij}$ = GWT of the image I of size (M×N) where i=0,…,4;j=0,…,7. are the scale and orientation variation.

**Step 2**: Find $I_{ij}$ = modulus of $G_{ij}$.

**Step 3**: Find $M_{ij}$ = mean of $I_{ij}$.

**Step 4**: Divide the matrix $I_{ij}$ into square blocks of size (W×W). Thus total no. of blocks = $\left\lfloor \frac{M}{W} \right\rfloor \times \left\lfloor \frac{N}{W} \right\rfloor$, denote it by $B_{ijk}$ where k=1,…, $\left\lfloor \frac{M}{W} \right\rfloor \times \left\lfloor \frac{N}{W} \right\rfloor$.

**Step 5**: Find vector $\chi_{ijk}$ such that it contains pixel values of $B_{ijk}$, which are greater than $M_{ij}$. If $\chi_{ijk} = [\ ]$, then put $\chi_{ijk} = M_{ij}$.

**Step 6**: For t=1 to length of ($\chi_{ijk}$)

if $|\chi_{ijkt} - \overline{\chi}_{ijk}| < threshold$,

take $Y_{ijk} = [\chi_{ijkt}]$, where $\overline{\chi}_{ijk}$ = mean of $\chi_{ijk}$.

if $Y_{ijk} = [\ ]$, then $Y_{ijk} = \overline{\chi}_{ijk}$. Threshold is taken as 3.

**Step 7**: $Z_{ijk}$ = Sort ($Y_{ijk}$) descending order of the values.

**Step 8**: L= $\underset{k}{Minimum}(length\ of\ (Z_{ijk}))$

**Step 9**: Select the top L elements of every vector $Z_{ijk}$ and store it in a column vector $\chi_{I_{ij}}$. Thus we get high intensity feature vector $\chi_{I_{ij}}$.

**Step 10**: Repeat step 2 to step 9 for all $I_{ij}$ of image I.

**end**

(b) Feature vector generation: The final high intensity feature vector $\chi_k$ is generated by accumulation of the elements of $(\chi_{I_{ij}})$ column wise to a single vector for the $k^{th}$ individual faces. The advantage of this feature extraction over, the previous algorithm [17] is that, previously the computation was dependent on the number of the GW responses, which is eliminated here. Thus the time complexity of this new algorithm is reduced by a factor of '$\lambda$' where $\lambda$ is the number of GW responses.

## 4. ICA on the extracted High-Intensity Feature Vectors of Gabor responses

As ICA would further reduce redundancy and represent independent features explicitly [20]. So independence property of these high-intensity features facilitates the application for image classification. These images, thus, produce salient local features that are most suitable for face recognition. On the other hand, we have input representations that are robust to noise and better reflect the data.

### 4.1 Preprocessing for ICA

Features of facial images are obtained through eight directions and five scales respectively. So for each face image, we get 5×8=40 Gabor transformed images. So dimension reduction becomes necessary for the high dimensional extracted important facial features. In this paper independent component analysis (ICA) is done using FastICA. Thus some preprocessing becomes necessary on the extracted feature vectors before applying the FastICA [30]: i) Centering: Here the mean vector m = $E\{x\}$ is subtracted from x to make x a zero-mean variable. This preprocessing is made solely to simplify the ICA algorithms. As the face features have large difference mean, so centering is needed before implementation of FastICA. ii) Whitening: Whitening is done by $\tilde{x} = ED^{-1/2}E^T x$ where the diagonal matrix D and the orthogonal matrix E are obtained by PCA of $xx^T$. Whitening is done because whitening reduces the number of parameters to be estimated.

### 4.2 Implementation of FastICA

ICA learns the higher-order dependencies in addition to the second-order dependencies among the pixels. PCA driven coding schemes are optimal and useful only with respect to data compression and decorrelation of lower (second) order statistics. The ICA method, which expands on PCA as it considers higher (>2) order statistics, is used here to derive independent high intensity features found useful for the recognition of human faces. Whitening reduces the number of parameters to be estimated. So instead of estimating $n^2$ parameters that are the elements of the original matrix **A**, we need to estimate, only the mixing matrix **A** [30]. An orthogonal matrix contains n(n−1)/2 degrees of freedom. Whitening is done to reduce the complexity of the problem i.e. to reduce the dimension of the data. This is done by discarding the eigen values $d_j$ of $E\{xx^T\}$ which are too small.[31]. This also reduces the effect of noise. Again, dimension reduction also prevents over-learning. In this paper, the data has been preprocessed by centering and whitening, before using FastICA. Here the FastICA algorithm has been chosen based on a fixed-point iteration scheme proposed by Hyvärinen and Oja [31], which was derived using an approximative Newton iteration. The algorithm is motivated as follows:

1. Choose an initial random weight vector w.
2. Let $w^+ = E\{xg(w^T x)\} - E\{g'(w^T x)\}w$
3. Let $w = w^+ / \|w^+\|$
4. If not converged, go back to 2.

The function $g(u)$ chosen here is the derivative of $f(u) = u^4$. Here it is here assumed that the data is prewhitened. The algorithm FastICA was introduced in two versions: i) a one-unit approach, and ii) a symmetric one. The first step, which is common for both versions and for many other ICA algorithms, consists in removing the sample mean the decorrelation of the data X, i.e., $Z = \hat{C}^{1/2}(X - \overline{X})$ where $\hat{C} = (X - \overline{X})(X - \overline{X})^T$ and $\overline{X}$ is the sample mean of the measured data. The derivation of FastICA is as follows. The maxima of the approximation of the negentropy of $w^T x$ are obtained at certain optima of $E\{G(w^T x)\}$. According to the Kuhn-Tucker conditions (shown by Hyvärinen, A. and Oja, E. (1997), [31]), the optimum of $E\{G(w^T x)\}$ under the given constraint $E\{G(w^T x)^2\} = \|w\|^2 = 1$ can be obtained at points where

$$E\{g(w^T x)\} - \beta w = 0 \quad (6)$$

Newton's method can be used to solve equation (6). The Jacobian matrix of the equation (6) can be written as $JF(w) = E\{xx^T g'(w^T x)\} - \beta I$.

Since the data is sphere, it can be approximated in the following way:

$$E\{xx^T g'(w^T x)\} \approx (xx^T) E(g'(w^T x)) = E(g'(w^T x))I$$

Thus the Jacobian matrix becomes diagonal, and can easily be inverted. Thus the following can be obtained with the approximative Newton iteration:

$$w^+ = w - [E\{xg(w^T x)\} - \beta w] / [E\{g'(w^T x)\} - \beta] \quad (7)$$

Usually the expectation of FastICA is replaced by their estimates. The natural estimates are generally the sample means. The one-unit ICA is based on minimization/maximization of the criterion $c(w) = E[G(x^T Z) - G_0]^2$ where w is the to-be found vector of coefficients that separates a desired independent components(ICs) from the mixture, E stands for the for the sample mean, $G(\cdot)$ is a suitable nonlinear function, called contrast function and $G_0$ is the expected value of $G(\eta)$ where $\eta$ is a standard normal random variable. The symmetric FastICA estimates all signals in parallel, and each step is completed by a symmetric orthogonalization:

$$W^+ \leftarrow g(WZ)Z^T - \text{diag}[g'(WZ)1_N]W$$
$$W \leftarrow (W^+ W^{+T})^{-1/2} W^+,$$

where $g(\cdot)$ and $g'(\cdot)$, denote the first and the second derivative of $G(\cdot)$, respectively. Hyvärinen [31], formulated the likelihood in the noise-free ICA model, and then estimate the model by a maximum likelihood method, which is denoted by W= $(w_1,..,w_n)^T$ the matrix $A^{-1}$, the log-likelihood takes the form

$$W^+ = W + \mu[1 + g(y)y^T]W \quad (8).$$

where μ is the learning rate.

Here, g is a function of the probability density function (pdf) of the independent components: $g = f_i' / f_i$ where $f_i$ is the pdf of an independent component. The column vectors of $W^+$ in (7) define a unique ICA representation. The feature base-vector of ICA is statistically independent. Thus it reflects the global feature of image and also local and edge features. Thus nature of image statistics can be more fully revealed by ICA. The IHIF method applies ICA on the high-intensity vectors obtained from (6). Lastly the IHIF method derives the overall ICA transformation matrix denoted by $W^+$. Next classification of the face images is done by nearest neighbor algorithm, using different classifiers.

## 5. Experiment

The Gabor-based ICA method integrates the Gabor wavelet representation of face images and ICA for face recognition. When a face image is presented to the Gabor-based ICA classifier, the low-dimensional high intensity Gabor feature vector of the image is first calculated as detailed in Section 2, then fastICA is applied on the low-dimensional most important high intensity Gabor feature vector to finally obtain the most important independent high-intensity feature vector, which is used as the input data instead of the whole image. Thus we have input representations that are robust to noise and better reflect the data.

Let $M_k^0$, k=1, 2..., l, be the mean of the training samples for class $\omega_k$. The classifier applies, then, the nearest neighbor (to the mean) rule for classification using some similarity measure $\delta$ :

$$\delta(\Im, M_k^0) = \min_j \delta(\Im, M_k^0) \to \Im \in \omega_k \quad (9).$$

The independent high-intensity feature vector, $\Im$ is classified as belonging to the class of the closest mean, $M_k^0$, using the similarity measure $\delta$. The similarity measure used here are, $L_2$ distance measure, $\delta_{L_2}$, and the cosine similarity measure, $\delta_{\cos}$, which are defined as:

$$\delta_{L_2} = (X - Y)^T (X - Y), \quad (10)$$

$$\delta_{\cos} = \frac{-X^T Y}{\|X\|\|Y\|}, \quad (11)$$

where $\|.\|$ denotes the norm operator, T denotes the transpose operator. Note that the cosine similarity measure includes a minus sign in (11) because the nearest neighbor (to the mean) rule of (9) applies minimum (distance) measure rather than maximum similarity measure.

### 5.1 Experiments of the Proposed method on Frontal and Pose-Angled Images for Face Recognition

This section assesses the performance of the Gabor-based ICA method for both frontal and pose-angled face recognition. The effectiveness of the Gabor-based ICA method is successfully tested on the ORL, FRAV2D and FERET face database. For frontal face recognition, the data set is taken from the ORL and the FRAV2D database. The dataset from ORL contains 400 frontal face images corresponding to 40 individuals, and the dataset from FRAV2D contains 1440 frontal face images corresponding to 120 individuals. The images are acquired under variable illumination, with occluded face features and facial expression. For pose-angled face recognition, the data set is taken from the FERET database, and it contains 1200 images with different facial expressions and poses of 120 individuals. Comparative performance of the proposed method is shown against the PCA method, the kernel PCA method. Performance results on FERET database has shown that proposed method is successive than the PCA, LDA and kernel PCA based algorithms.

#### 5.1.1 ORL Database

The ORL database, employed here consists of 10 frontal face images of each individual consisting a total of 40 individual with different facial expressions and head pose (tilting and rotation up to 20 degrees), illumination condition also has slightly changes. Here each image is scaled to 92 × 112 with 256 gray levels. Fig. 3 shows all samples of one individual. Five images are randomly chosen from each individual of the 40 individual for training samples and the rest 5 images are used as testing samples.

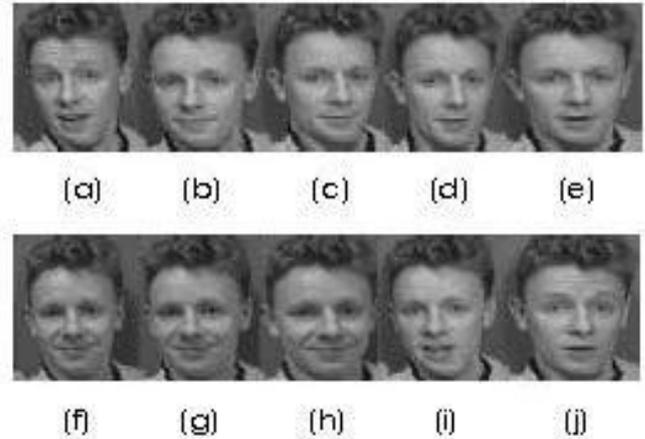

**Fig. 3.** Demonstration images of one individual from the ORL Database

#### 5.1.2 FRAV2D Face Database

The FRAV2D face database, employed in the experiment consists; 1320 colour face images of 120 individuals, 12 images of each individual are taken, including frontal views of faces with different facial expressions, under different lighting conditions. All color images are transformed into gray images and are scaled to 92×112. Randomly five images of each individual were chosen i.e., is a total of 600 images as training samples and the remaining 840 images are regarded as testing samples. Fig. 4 show all samples of one individual.

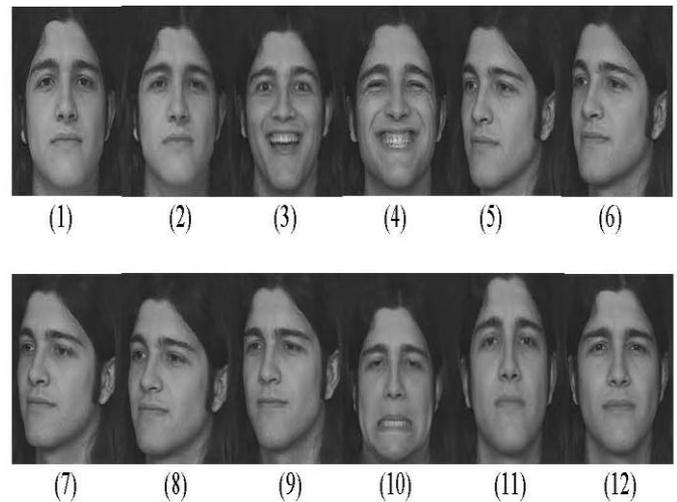

**Fig. 4**. Demonstration images of one individual from the FRAV2D database

#### 5.1.3 FERET Face Database

The FERET database, employed in the experiment here contains, 1,200 facial images corresponding to 120 individuals with each individual contributing 10 images. The images in this database were captured under various illuminations and display a variety of facial expressions and poses. As the images include the background and the body chest region, so each image is manually cropped to exclude those, and finally scaled to 92×112. Fig. 5 shows all samples of one individual. Randomly five images from each individual were chosen i.e., is a total of 600 images are considered as training samples and the remaining 600 images are regarded as testing samples.

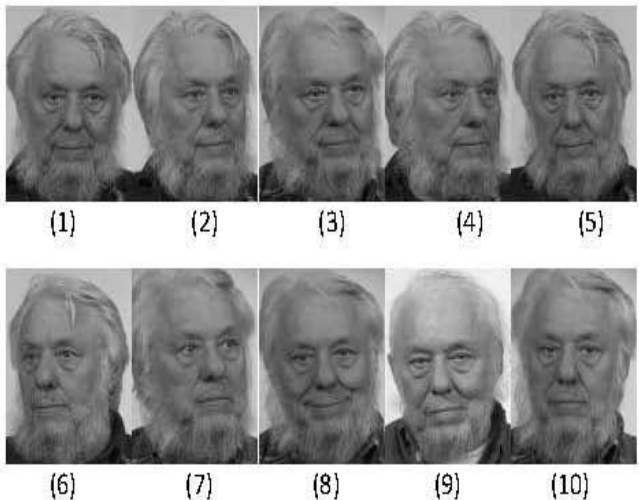

**Fig. 5.** Demonstration images of one individual from the FERET database.

### 5.2 Specificity and Sensitivity

Sensitivity and specificity [32] of the performance of the proposed method are measured on the dataset from the FERET, FRAV2D and ORL databases. In this regard, the true positive rate $T_P$; false positive rate $F_P$; true negative rate $T_N$; and false negative rate $F_N$: are also being measured. To measure the sensitivity and specificity, the dataset from the FERET database consists of a total of 100 class, in which each class have a total of 15 images, of which 10 images are of a particular individual, (in which 5 images are randomly used for training, and the left 5 images of the particular individual are used for positive testing), and the rest 5 images of other individuals are considered for negative testing. Fig. 6 shows all sample images of one class of the data set used from FERET database.

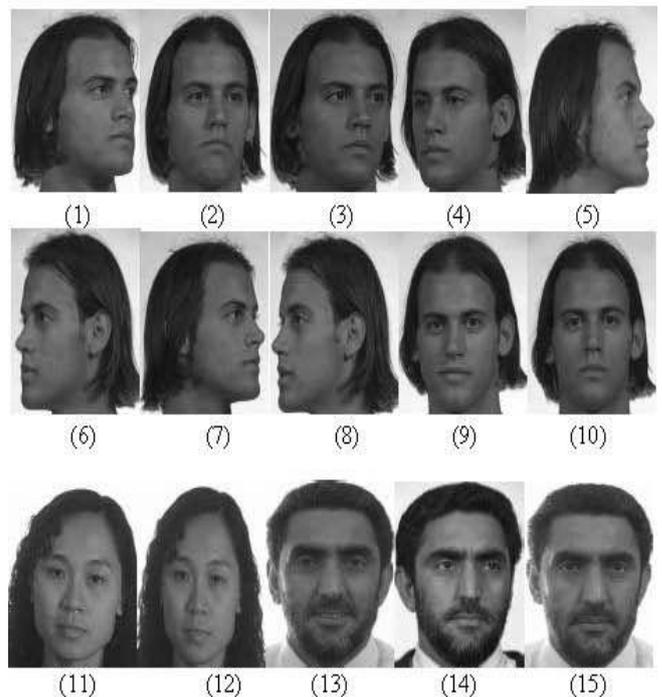

**Fig. 6**. Demonstration images of one class from the FERET database.

The dataset taken from the FRAV2D database consists of 120 classes, in which each class contains a total of 18 images, out of which 12 images are of particular individual, (in which 5 images are randomly used for training, remaining 7 images of the particular individual were used for positive testing), and the rest 6 images of other individuals were considered for negative testing. Fig. 7 shows all sample images of one class of the dataset used from FRAV2D database used to measure the sensitivity and specificity.

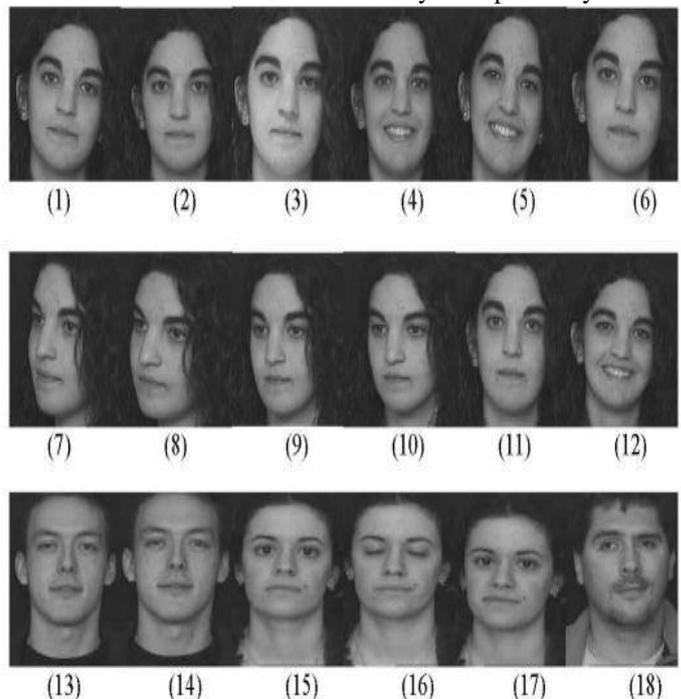

**Fig. 7.** Demonstration images of one class from the FRAV2D database

The dataset taken from the ORL database consists of 40 classes, in which each class contains a total of 15 images, out of which 10 images are of a particular individual, (in which 5 images are randomly used for training, remaining 5 were used for positive testing), and the remaining 5 images of other individuals are taken for negative testing. Fig. 8 shows all sample images of one class of the dataset used from ORL database used to measure the sensitivity and specificity.

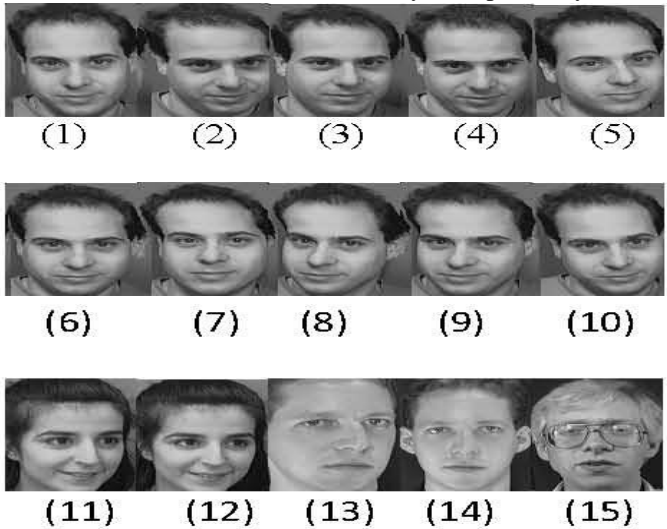

**Fig. 8.** Demonstration images of one class from the ORL database.

### 5.3 Experimental Results

Experiments were conducted that implements Gabor based ICA method, with different similarity measures on the ORL, FERET and FRAV2D database to measure both the positive and negative recognition rates i.e. $T_P$; $F_P$; $T_N$; $F_N$ and hence the sensitivity and specificity. In this paper, we use only the high intensity features of the Gabor transformed image to include in ICA defined by (8). In order to derive the independent high-intensity ECA features. Fig. 9 and Fig. 10 shows face recognition performance of the IHIF method in terms of specificity and sensitivity respectively using the two different similarity measures. In our experiment, the algorithm would attempt to separate 500 ICs. Although it is shown already [12] that performance improves with the number of components separated. But, 1000 becomes a little bit intractable, so in order to have control over the number of ICs extracted by the algorithm, instead of performing ICA on the original images, ICA was applied on the set of extracted high-intensity feature vector obtained from the GW responses of the image. In Fig. 9 the horizontal axis indicates the number of ICs used, and the vertical axis represents the specificity rate of the face recognition, thus measures the proportion of negatives which are correctly identified which is the true negative rate i.e. correct rejection. The top response is the correct rejection rate. In Fig. 10 the horizontal axis indicates the number of ICs used, and the vertical axis represents the sensitivity rate of the face recognition, thus measures the proportion of actual positives which are correctly identified as such. From both the measure it is seen that the cosine similarity distance measure performs the best. This shows that cosine similarity distance measure further enhances face recognition.

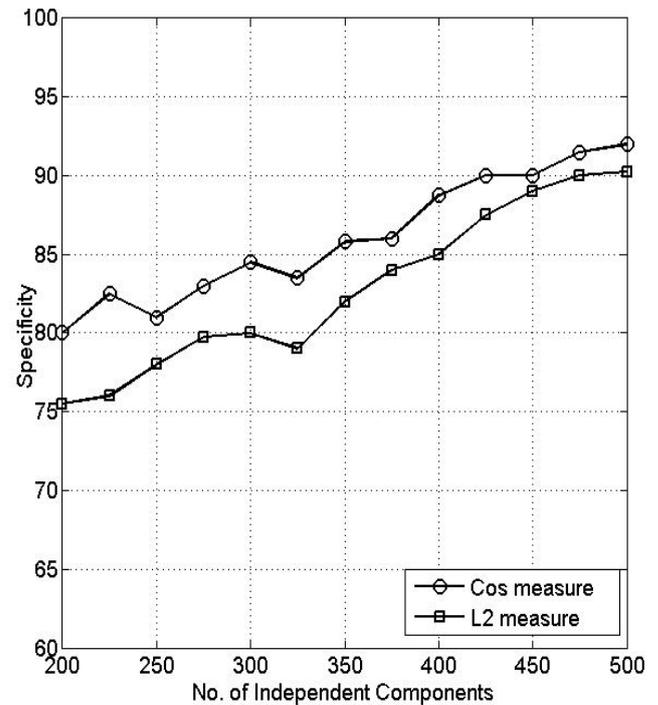

**Fig. 9.** Face recognition performance of the IHIF method on the FERET database, using the $L_2$ (the $L_2$ distance measure), and cos (the cosine similarity measure) for measuring negative recognition accuracy.

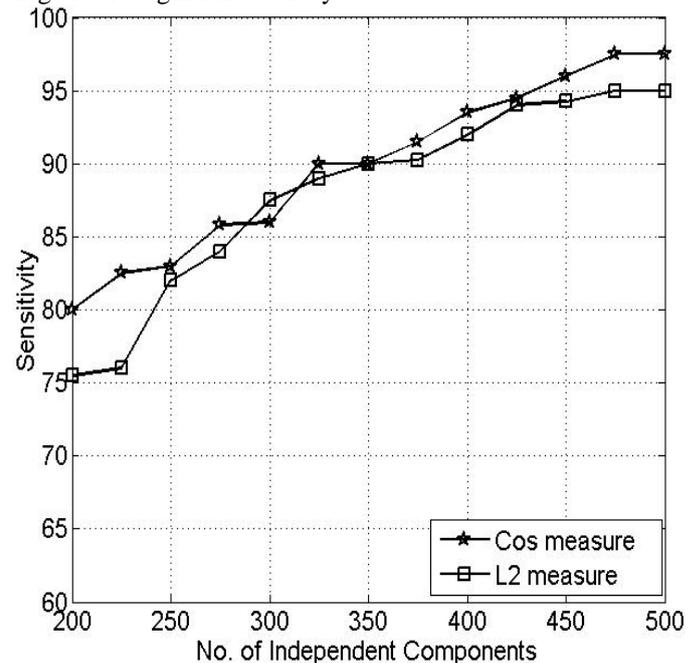

**Fig.10**. Recognition performance of the IHIF method on the FERET database, using the similarity measures $L_2$ (the $L_2$ distance measure), and cos (the cosine similarity measure) for measuring positive recognition accuracy.

As the proposed method performs best with the cosine similarity distance classifier, so experiments were conducted on the ORL, FERET and FRAV2D database to assess the face recognition performance in terms of specificity and sensitivity of the IHIF method, using the cosine similarity distance measure are shown in Fig.11 and Fig. 12.

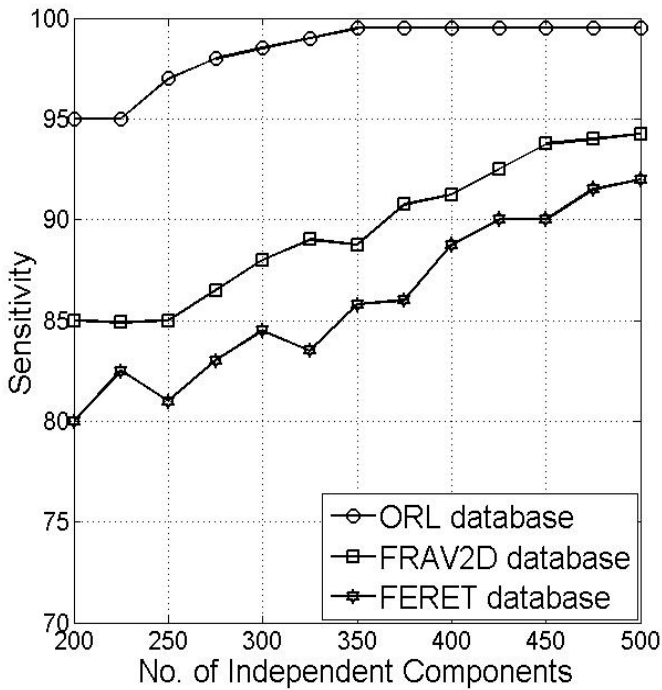

**Fig. 11**. Positive recognition performance of the IHIF method using the cosine similarity distance measure on the ORL, the FRAV2D and the FERET database.

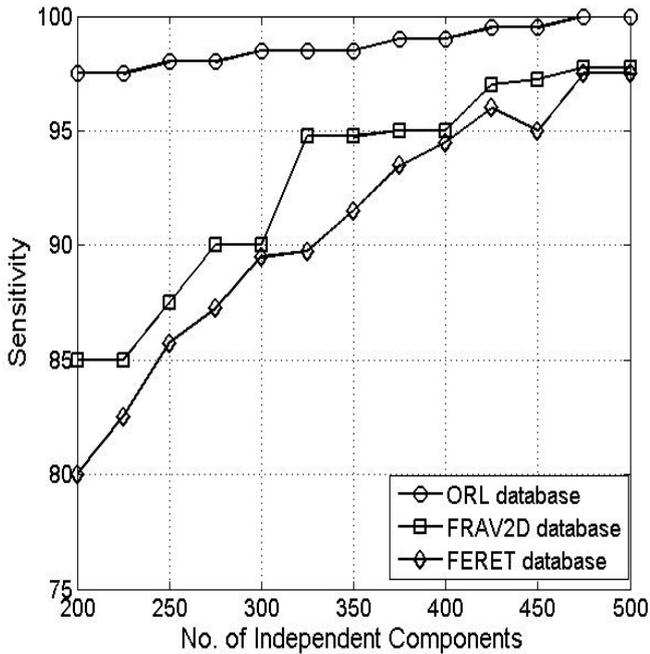

**Fig. 12.** Negative recognition performance of the proposed method using cosine similarity distance measure on the ORL, FRAV2D and the FERET database.

Specificity and Sensitivity measure of the proposed method for face recognition using the three databases, a) the FERET facial database, b) the FRAV2D facial database and the c) ORL database are shown in table1 table2 and table 3 respectively.

**Table 1.** The dataset from FERET database consists of 1200 images of 120 individuals,( in which each class onsists of 10 images of one individual and 5 images of any other individual/individuals) taken by permutation:

|  |  | Individual belonging to a particular class | |
|---|---|---|---|
|  |  | Positive | Negative |
| FERET test | Positive | $T_P = 552$ | $F_P = 12$ |
|  | Negative | $F_N = 48$ | $T_N = 488$ |
|  |  | Sensitivity = $T_P / (T_P + F_N)$ = 92% | Specificity = $T_N / (F_P + T_N)$ ≈ 97.6% |

False positive rate = $F_P / [F_P + T_N]$ = 1 − Specificity = 2.9%
False negative rate = $F_N / (T_P + F_N)$ = 1 − Sensitivity = 8.0%
Accuracy = $(T_P+T_N)/(T_P+T_N+F_P+F_N)$ = 94.8

**Table 2:** The data set from FRAV2D database consists of
1200 images of 100 individuals (in which each class folder consists of 12 images of one individual and 6 of any other individual/individuals) taken by permutation:

|  |  | Individual belonging to a particular class | |
|---|---|---|---|
|  |  | Positive | Negative |
| FRAV2D test | Positive | $(T_P) = 797$ | $(F_P) = 14$ |
|  | Negative | $(F_N) = 43$ | $(T_N) = 586$ |
|  |  | Sensitivity = $T_P / (T_P + F_N)$ ≈ 94.85% | Specificity = $T_N / (F_P + T_N)$ = 97.8% |

**False positive rate** = $F_P / (F_P + T_N)$ = 1 − Specificity = **2.2%**
**False negative rate** = $F_N / (T_P + F_N)$ = 1 − Sensitivity = **5.15%**
**Accuracy** = $(T_P+T_N)/(T_P+T_N+F_P+F_N)$ ≈ **96.32**.

**Table 3:** The dataset from ORL database consists of 400 images of 40 individuals ( in which each class consists of 10 images of one individual and 5 image of any other individual/
/ individuals) taken by permutation:

|  |  | Individual belonging to a particular class | |
|---|---|---|---|
|  |  | Positive | Negative |
| ORL test | Positive | $(T_P) = 199$ | $(F_P) = 0$ |
|  | Negative | $(F_N) = 01$ | $(T_N) = 200$ |
|  |  | Sensitivity = $T_P / (T_P + F_N)$ ≈ 99.5% | Specificity = $T_N / (F_P + T_N)$ = 100% |

**False positive rate** = $F_P / (F_P + T_N)$ = 1 − Specificity = **0%**
**False negative rate** = $F_N / (T_P + F_N)$ = 1 − Sensitivity = **.5%**
**Accuracy** = $(T_P+T_N)/(T_P+T_N+F_P+F_N)$ = **99.75.**

Recognition performances on ORL database of well-known methods are cited in previous works as: Eigenfaces 80.0% [33], Elastic matching 80.0% [33]. Neural network 96.2% [34], line based 97.7% min [35]. Although, reported recognition rates are higher in convolutional neural network and line-based method. It must be noted that these two are using more than one facial image for each individual in training. Our method achieves using only one reference facial image for each individual. So, the proposed method **gets** better results compared to similar methods. Tests on FERET database are held in accordance with the FERET procedure [36], and shown in table 4.

**Table 4.** Performance results of well-known algorithms and IHIF method on FERET database.

| Method | Recognition Rate (%) |
|---|---|
| Eigenface method with Bayesian Similarity measure | 79.0 |
| Elastic graph matching | 84.0 |
| A Linear Discriminant Analysis based algorithm | 88.0 |
| Kernel PCA | 91.0 |
| Line based | 92.7 |
| Neural network | 93.5 |
| **Proposed IHIF Method with Cos measure** | **92** |

The images considered here, consists of frontal faces with different facial expressions, illumination conditions, and occlusions. To examine the robustness of the proposed method to the illumination changes, the IHIF method has been experimented especially on FERET database. Experimental results indicate that a) the extracted high-intensity feature points by the proposed method enhance the face recognition performance in presence of occlusions as well as reduce computational complexity compared to the EGM [6]. b) The ICA applied on the extracted high-intensity feature vectors further enhances recognition performance. Our results show that 1) the IHIF method greatly enhance the recognition performance, and also reduces the dimensionality of the feature space when compared only with high-intensity features. 2) The cosine similarity measure classifier further enhances the recognition performance

## 6. Conclusion

In this paper a new approach to face recognition has been proposed by selecting the high-intensity features from the GWT facial images. Then ICA is applied on these extracted high-intensity feature vectors to obtain the IHIF vectors. As ICA employ local salient information, so here the extracted facial features are compared locally instead of a general structure, hence it allows us to make a decision from the different parts of a face, and thus maximizes the benefit of applying the idea of "recognition by parts". Thus it performs well in presence of occlusions (sunglasses scarf etc.), this is due to the fact that when there are sunglasses or any other obstacles the algorithm compares face in terms of mouth, nose and other features rather than the eyes. The algorithm also reduces computational cost as ICA is applied, only on the few extracted high-intensity features vectors of the GWT image instead of the whole image. Moreover, the proposed method has a simple matching procedure, and is also robust to illumination changes, as a property of GW's, together with the application of ICA. From the experimental results it is seen that proposed method achieves better results compared to other well known algorithms [33], [34], [35], [36] which are known to be the most successful algorithms. Experimental results also reveal that, ICA performs better using the cosines similarity distance measure than the Euclidean distance ($L_2$) measure.

## Acknowledgement

Authors are thankful to the "Centre for Microprocessor Application for Training Education and Research" at the Department of Computer Science and Engineering, Jadavpur University, Kolkata - 700 032 for providing the necessary facilities for carrying out this work. Dr. D.K. Basu acknowledges with thanks the receipts of AICTE Emeritus Fellowship (1-51/RID/EF(13)/2007-08).